\documentclass[article]{IEEEtran}
\usepackage[utf8]{inputenc}
\usepackage{graphicx} 
\usepackage{multirow}
\usepackage{multicol} 
\usepackage{booktabs}
\usepackage{amssymb}
\usepackage{pifont}
\usepackage{gensymb}
\usepackage{nopageno} 
\usepackage{caption}
\usepackage{subcaption}
\usepackage{algorithm}
\usepackage{algpseudocode}
\usepackage{bm}
\usepackage{amsmath}
\usepackage{mathtools}
\usepackage{commath}
\usepackage{nomencl}

\usepackage{tikz}
\usepackage{pgf}
\usepackage{pgfplots}
\usepackage{tikz-network}
\usepackage{authblk}

\usepackage{float}
\newfloat{algorithm}{t}{lop}

\graphicspath{{Figures/}}

\DeclareMathSymbol{\ThetaCal}{\mathord}{operators}{"02}
\algnewcommand{\Initialize}[1]{%
  \State \textbf{Initialize}
  \Statex \hspace*{\algorithmicindent}\parbox[t]{.8\linewidth}{\raggedright #1}
}
\algnewcommand{\Solve}[1]{%
  \State \textbf{Solve}
  \Statex \hspace*{\algorithmicindent}\parbox[t]{.8\linewidth}{\raggedright #1}
}
\algnewcommand{\RequireR}[1]{%
  \State \textbf{Require}
  \Statex \hspace*{\algorithmicindent}\parbox[t]{.8\linewidth}{\raggedright #1}
}

\renewcommand{\footnoterule}{%
\kern -3pt
\hrule width 2in
\kern 2.6pt
}

\usepackage{fixltx2e}

\usepackage{graphicx}

\usepackage{subcaption}              

\usepackage[utf8]{inputenc}

\usepackage{amsfonts}
\usepackage{amsmath}
\usepackage{mathtools}

\usepackage{amssymb}

\usepackage{bm}

\usepackage{color,verbatim}
\usepackage{multirow}
\usepackage{accents}

\usepackage{theoremref}

\usepackage{flushend}

\usepackage{url}

\newcounter{rulecounter}
\newcommand{\resetrule}{ \setcounter{rulecounter}{0}}
\resetrule

\newsavebox{\selvestebox}
\newenvironment{colbox}[1]
  {\newcommand\colboxcolor{#1}%
   \begin{lrbox}{\selvestebox}%
   \begin{minipage}{\dimexpr\columnwidth-2\fboxsep\relax}}
  {\end{minipage}\end{lrbox}%
   \begin{center}
   \colorbox{\colboxcolor}{\usebox{\selvestebox}}
   \end{center}}

\definecolor{orange}{rgb}{1,0.8,0}
\definecolor{gray}{rgb}{.9,0.9,0.9}
\definecolor{darkgray}{rgb}{.3,0.3,0.3}
\definecolor{darkblue}{rgb}{.1,0.0,0.3}
\definecolor{lightblue}{rgb}{0.7,0.7,1}
\definecolor{lightred}{rgb}{1,0.7,.7}
\definecolor{purple}{RGB}{204,153,255}
\definecolor{lightgray}{rgb}{.95,0.95,0.95}
\definecolor{lightgreen}{rgb}{0.3,0.5,0.3}
\definecolor{darkgreen}{rgb}{0.05,0.3,0.05}

\newtheorem{myproposition}{Proposition}
\newtheorem{myremark}{Remark}
\newtheorem{myproblemstatement}{Problem Statement}
\newtheorem{mylemma}{Lemma}
\newtheorem{mytheorem}{Theorem}
\newtheorem{mydefinition}{Definition}
\newtheorem{mycorollary}{Corollary}

\usepackage{pgfplots}
\pgfplotsset{compat=newest}
\usetikzlibrary{plotmarks}
\usetikzlibrary{positioning}
\usetikzlibrary{arrows.meta}
\usepgfplotslibrary{patchplots}
\usepackage{grffile}

\pgfplotsset{plot coordinates/math parser=false}
\newlength\mywidth
\newlength\myheight
\newlength\mywidths
\newlength\myheights
\newlength\mywidthss
\newlength\myheightss
\definecolor{mycolor1}{rgb}{0.00000,0.44700,0.74100}%
\definecolor{mycolor2}{rgb}{0.85000,0.32500,0.9800}%
\definecolor{mycolor3}{rgb}{0.92900,0.69400,0.12500}%
\definecolor{mycolor4}{rgb}{0.89400,0.18400,0.15600}%
\definecolor{mycolor5}{rgb}{0.46600,0.67400,0.18800}%
\definecolor{mycolor6}{rgb}{0.30100,0.74500,0.93300}%
\definecolor{mycolor7}{rgb}{0.63500,0.07800,0.18400}%

\definecolor{AGNN}{rgb}{1,0.04700,0.04100}%
\definecolor{GCN}{rgb}{0.05000,0.32500,0.89800}%
\definecolor{AGNNnf}{rgb}{1,0.8,0.119800}%
\definecolor{Mune}{rgb}{0.2,0.800,1}%

\definecolor{TightHann}{rgb}{0.92900,0.69400,0.12500}%
\definecolor{pDiffScater}{rgb}{0.49400,0.18400,0.55600}%
\definecolor{pMonicCubic}{rgb}{0.46600,0.67400,0.18800}%
\definecolor{pTightHann}{rgb}{0.30100,0.74500,0.93300}%

\pgfplotscreateplotcyclelist{colorlist}{%
color=AGNN,line width=2.0pt,
solid, every mark/.append style={solid, fill=AGNN}, mark=*\\%
color=GCN,line width=2.0pt,dotted, every mark/.append style={solid, fill=GCN}, mark=square*\\%
color=AGNNnf,line width=2.0pt,densely dotted, every mark/.append style={solid, fill=AGNNnf}, mark=otimes*\\%
color=Mune,line width=2.0pt,loosely dotted, every mark/.append style={solid, fill=Mune}, mark=triangle*\\%
color=mycolor5,line width=2.0pt,
dashed, every mark/.append style={solid, fill=mycolor5},mark=diamond*\\%
color=mycolor6,line width=2.0pt,loosely dashed, every mark/.append style={solid, fill=mycolor6},mark=*\\%
color=mycolor7,densely dashed, every mark/.append style={solid, fill=mycolor7},line width=2.0pt,mark=square*\\%
dashdotted, every mark/.append style={solid, fill=mycolor7},mark=otimes*\\%
dashdotdotted, every mark/.append style={solid},mark=star\\%
densely dashdotted,every mark/.append style={solid, fill=gray},mark=diamond*\\%
}
\pgfplotscreateplotcyclelist{my black white}{%
solid, every mark/.append style={solid, fill=gray}, mark=*\\%
dotted, every mark/.append style={solid, fill=gray}, mark=square*\\%
densely dotted, every mark/.append style={solid, fill=gray}, mark=otimes*\\%
loosely dotted, every mark/.append style={solid, fill=gray}, mark=triangle*\\%
dashed, every mark/.append style={solid, fill=gray},mark=diamond*\\%
loosely dashed, every mark/.append style={solid, fill=gray},mark=*\\%
densely dashed, every mark/.append style={solid, fill=gray},mark=square*\\%
dashdotted, every mark/.append style={solid, fill=gray},mark=otimes*\\%
dashdotdotted, every mark/.append style={solid},mark=star\\%
densely dashdotted,every mark/.append style={solid, fill=gray},mark=diamond*\\%
}
\usetikzlibrary{decorations.pathreplacing}
\definecolor{lavander}{cmyk}{0,0.48,0,0}
\definecolor{violet}{cmyk}{0.79,0.88,0,0}
\definecolor{burntorange}{cmyk}{0,0.52,1,0}
\definecolor{mygreen}{rgb}{0,1,0}%
\definecolor{myred}{rgb}{1,0,0}%

\tikzstyle{nnblock}=[draw,rectangle,  rounded corners,text=black,
minimum width=30pt,minimum height=60pt]
\tikzstyle{esl}=[draw,rectangle,  rounded corners,text=black,
minimum width=60pt,minimum height=30pt]
\tikzstyle{grnn}=[draw,rectangle,  rounded corners,text=black,
minimum width=30pt,minimum height=15pt]
\tikzstyle{graphsampler}=[draw,rectangle,  fill=orange,fill opacity=0.1, rounded corners,text=orange,
minimum width=30pt,minimum height=15pt]
\tikzstyle{outblock}=[draw,rectangle, color=red, rounded corners,text=black,
minimum width=10pt,minimum height=60pt]
\tikzstyle{nam}=[rectangle, fill=orange,fill opacity=0.1, rounded corners,text=orange,
minimum width=8pt,minimum height=58pt]
\tikzstyle{gam}=[fill=blue,fill opacity=0.1, rectangle,  rounded corners,text=black,
minimum width=8pt,minimum height=58pt]
\tikzstyle{fam}=[fill=green,fill opacity=0.1, rectangle,  rounded corners,text=black,
minimum width=8pt,minimum height=58pt]

\tikzstyle{peers}=[draw,circle,bottom color=myred,
                  top color= white, text=violet,minimum width=8pt]
\tikzstyle{superpeers}=[draw,circle, left color=mygreen,
                       text=violet,minimum width=8pt]

\tikzstyle{peers3}=[draw,circle,bottom color=myred,
                  top color= white, text=violet,minimum width=12pt]
\tikzstyle{superpeers3}=[draw,circle, left color=mygreen,
                       text=violet,minimum width=12pt]
\tikzstyle{peers2}=[draw,circle,bottom color=myred,
                  top color= white, text=violet,inner sep=0pt,minimum size=0.1cm]
\tikzstyle{superpeers2}=[draw,circle, left color=mygreen,
                       text=violet,inner sep=0pt,minimum size=0.1cm]
                    
\tikzstyle{peers1}=[draw,circle,bottom color=myred,
                  top color= white, text=violet,minimum width=4pt]
\tikzstyle{superpeers1}=[draw,circle, left color=mygreen,
                       text=violet,minimum width=4pt]
\tikzstyle{legendsp}=[rectangle, draw,  rounded corners,
                     thin,bottom color=mygreen, top color=white,
                     text=black, minimum width=2cm]
\tikzstyle{legendp}=[rectangle, draw,  rounded corners, thin,
                     bottom color=myred, top color= white,
                     text= black, minimum width= 2cm]
\tikzstyle{legend_general}=[rectangle, rounded corners, thin,
                           burntorange, fill= white, draw, text=violet,
                           minimum width=2.5cm, minimum height=0.8cm]
\usetikzlibrary{shapes.misc, positioning}

\usetikzlibrary{calc,trees}

\title{3D Scene Rendering with Multimodal Gaussian Splatting
\\
	\thanks{\;\; This work was supported by the Eric and Wendy Schmidt AI for Science, the NSF TILOS AI Institute, the UCSD Centers for Machine intelligence, computing, and security (MICS) and Wireless Communications (CWC), the ONR Award N00014-22-1-2363 and the NSF grant 2148313, with the latter being supported in part by funds from federal agency and industry partners as specified in the Resilient \& Intelligent NextG Systems (RINGS) program. } 
	\thanks{\;\;Emails: \textit{cgau@ucsd.edu, kpolyzos@ucsd.edu, abacharis@nvidia.com, vmadhuvarasu@ucsd.edu,  tjavidi@ucsd.edu}
}}

\author{Chi-Shiang Gau$^1$}
\author{Konstantinos D. Polyzos$^1$}
\author{Athanasios Bacharis$^2$}
\author{Saketh Madhuvarasu$^1$}
\author{Tara Javidi$^1$}
\affil{Department of Electrical and Computer Engineering, University of California San Diego, USA$^1$ \\
NVIDIA$^2$}

\begin{document}
\maketitle
\begin{abstract}

3D scene reconstruction and rendering are core tasks in computer vision, with applications spanning industrial monitoring, robotics, and autonomous driving. Recent advances in 3D Gaussian Splatting (GS) and its variants have achieved impressive rendering fidelity while maintaining high computational and memory efficiency. However, conventional vision-based GS pipelines typically rely on a sufficient number of camera views to initialize the Gaussian primitives and train their parameters, typically incurring additional processing cost during initialization while falling short  in conditions where visual cues are unreliable, such as adverse weather, low illumination, or partial occlusions. To cope with these challenges, and motivated by the robustness of radio-frequency (RF) signals to weather, lighting, and occlusions, we introduce a multimodal framework that integrates RF sensing, such as automotive radar, with GS-based rendering as a more efficient and robust alternative to vision-only GS rendering. The proposed approach enables efficient depth prediction from only sparse RF-based depth measurements, yielding a high-quality 3D point cloud for initializing Gaussian functions across diverse GS architectures. Numerical tests demonstrate the merits of judiciously incorporating RF sensing into GS pipelines, achieving high-fidelity 3D scene rendering driven by RF-informed structural accuracy.


\end{abstract}


\thispagestyle{empty} 

\section{Introduction}\label{sec:intro}

Lying at the crossroads of computer vision and robotics, 3D scene reconstruction and the ability to generate accurate 2D estimates from novel/unseen viewpoints, collectively referred to as 3D scene rendering, have become fundamental tasks due to their wide-ranging applications in autonomous driving, robotics, and surveillance, among others. Early remarkable advances in this domain were driven by neural radiance field (NeRF) methods \cite{mildenhall2021nerf, barron2021mip}, which demonstrated remarkable reconstruction and rendering fidelity but at the cost of substantial computational and memory demands. More recently, 3D Gaussian Splatting (GS) \cite{kerbl20233d} and its subsequent extensions \cite{liu2025review, bao20253d} have emerged as efficient and lightweight alternatives. By representing the 3D scene with a set of anisotropic Gaussian functions, GS achieves high-quality rendering while significantly reducing computational and memory overhead.

Given a fixed budget of available training camera views, the original and subsequent GS architectures rely on these images to (i) initialize the Gaussian primitives, typically by predicting a 3D point cloud (PC) to ensure proper alignment with the underlying 3D scene structure; and (ii) optimize the Gaussian parameters so that the rendered outputs match the ground-truth training images. Nonetheless, GS pipelines that depend on a large number of training views often incur substantial pre-processing overhead to generate this initial 3D PC, commonly via the traditional `structure-from-motion' process \cite{schoenberger2016sfm} or through pre-trained depth and 2D-3D correspondence models \cite{wang2024dust3r, leroy2024grounding}. To mitigate this burden and reduce redundancy, view-planning methods in robotics and active vision aim to identify a compact set of camera viewpoints for efficient PC generation; see, e.g., \cite{bacharis2025efficient, bacharis2025bosfm}. Motivated by the same objective in the context of GS and 3D rendering, the recently proposed ActiveInitSplat framework \cite{polyzos2025activeinitsplat} introduced an efficient active-view-selection strategy that identifies a small yet informative set of images for GS initialization and training, while being compatible with diverse GS architectures. Although effective and significantly more computationally efficient than passive GS pipelines, ActiveInitSplat still requires a non-negligible time to collect and process the actively selected views. More critically, ActiveInitSplat along with all other vision-only GS methods, is vulnerable in challenging real-world conditions where visual sensing degrades, including adverse weather conditions, low illumination, reduced image resolution, or partial occlusions.

While visual sensing can degrade under adverse conditions, complementary sensing modalities that use radio-frequency (RF) signals, such as automotive radar, offer robust alternatives for predicting depth and generating the corresponding 3D PC for reliable 3D reconstruction and rendering. In particular, RF signals exhibit strong robustness to weather, lighting conditions, and partial occlusions, making them well-suited for depth prediction when visual information becomes unreliable; see e.g., \cite{paek2022k, cui2023radar}. Even in scenarios where vision sensing quality is not degraded, obtaining a high-quality 3D PC from images often incurs a non-negligible runtime, restricting the practicality of such vision-only pipelines in real-time applications. 

Driven by these insights, we introduce a multimodal approach for efficient 3D scene rendering whose contributions can be summarized in the following aspects:
\begin{enumerate}
    \item[\textit{C1.}] We introduce an efficient RF-based depth prediction module that serves as a time- and computationally-efficient alternative to vision-based approaches for generating a reliable 3D PC for GS, while remaining robust under adverse conditions where visual cues become unreliable.  
    \item[\textit{C2.}] Using only sparse RF-based depth measurements, we introduce an efficient depth-map reconstruction approach that adapts conventional Gaussian Processes (GPs) through a principled localization scheme. By modeling different spatial regions with distinct local GPs, the proposed method provides more detailed uncertainty estimates and improves both computational efficiency and prediction accuracy at unobserved locations.
    \item[\textit{C3.}] Numerical tests on a real-world setting demonstrate the effectiveness of the proposed approach in combining RF and vision sensing modalities for efficient GS-based rendering.
\end{enumerate}

\section{Preliminaries and problem formulation}

GS has attracted significant attention from the research community due to its efficient representation of 3D scenes using a set of $N$ anisotropic Gaussian functions $\{G_i\}_{i=1}^N$ \cite{kerbl20233d}. To initialize these Gaussian functions, the original GS formulation and subsequent variants typically rely on a given budget of $T_{\text{train}}$ training images or camera views $\{\mathbf{I}_m\}_{m=1}^{T_{\text{train}}}$, to generate a set of 3D points called 3D point cloud (PC) of the scene of interest. This PC is commonly obtained through a structure-from-motion (SfM) pipeline \cite{schoenberger2016sfm} or estimated using pre-trained depth and 2D–3D correspondence models such as \cite{leroy2024grounding, wang2024dust3r}. 

With the PC at hand, each point $\mathbf{p}_i, i \in \{1,\ldots, N \}$ in the PC is associated with an anisotropic Gaussian function defined as 
\begin{align}
G_i(\mathbf{z}) = \alpha_i \text{exp}(-\frac{1}{2}(\mathbf{z}-\mathbf{m}_i)^\top \mathbf{\Sigma}_i^{-1} (\mathbf{z}-\mathbf{m}_i)) \label{eq:Gaus_def}
\end{align}
where $\mathbf{z}$ denotes any location in the 3D space at which the Gaussian is evaluated, $\mathbf{m}_i$ is the mean of $G_i$, initially placed at the position of the PC point $\mathbf{p}_i$, and $\mathbf{\Sigma}_i$ is the $3\times 3$ covariance matrix that determines the shape, size and orientation of $G_i$. The opacity parameter $\alpha_i$ controls the contribution of $G_i$ to the final rendered images.

Given these Gaussian functions, the color at any pixel $\mathbf{p}$ of the rendered 2D image is computed as \cite{kerbl20233d}
\begin{align}
\mathbf{c}(\mathbf{p}) = \sum_{i=1}^N \mathbf{c}_i G_i^{2D}(\mathbf{p}) \prod_{j=1}^{i-1} \left(1 - G_j^{2D}(\mathbf{p})\right) \label{eq:rend_formula}
\end{align}
where $\mathbf{c}(\mathbf{p})$ denotes the color at pixel $\mathbf{p}$, and $\mathbf{c}_i$ is the color associated with the $i$th Gaussian, parameterized either directly in RGB or as spherical harmonics (SH) to model view-dependent appearance. The term $G_i^{2D}(\cdot)$ denotes the 2D projection of the corresponding 3D Gaussian function $G_i$, whose mean and covariance are given by

\begin{subequations}
	\label{eq:proj_Gauss}
	\begin{align}
		\mathbf{m}_i^{2D} &= [u/b, l/b]^\top, \; [u,l,b]^\top = \mathbf{L} \mathbf{W} [\mathbf{m}_i,1]^\top
  \\	
		\mathbf{\Sigma}_i^{2D} &= \mathbf{J}\mathbf{W}\mathbf{\Sigma}_i\mathbf{W}^\top\mathbf{J}^\top
	\end{align}
\end{subequations}
where matrices $\mathbf{W}, \mathbf{L}$ denote the extrinsic and intrinsic camera parameter matrices, respectively, and $\mathbf{J}$ is the corresponding Jacobian; additional details are available in \cite{kerbl20233d}. 

All Gaussian parameters are optimized so that the rendered images match the ground-truth training views $\{\mathbf{I}_m\}_{m=1}^M$. To guarantee that each covariance matrix $\mathbf{\Sigma}_i$ remains positive semi-definite during optimization, it is parameterized using the factorization $\mathbf{\Sigma}_i = \mathbf{R}_i \mathbf{S}_i \mathbf{S}_i^\top \mathbf{R}_i^\top$ where $\mathbf{S}_i$ is the $3\times3$ diagonal scaling matrix and $\mathbf{R}_i$ is the rotation matrix represented analytically via quaternions.

In the vision domain, obtaining a high-quality PC that faithfully reflects the underlying 3D structure from training images $\{\mathbf{I}_m\}_{m=1}^{T_{\text{train}}}$ often requires (i) a sufficiently large number $T_{\text{train}}$ of views, whose processing, whether through SfM pipeline or pre-trained models, can be computationally expensive and unsuitable for time-critical applications; and (ii) images with adequate resolution and quality. These limitations naturally motivate the integration of alternative sensing modalities that are more time-efficient and remain robust under adverse conditions, such as weather or illumination ones, that degrade visual cues and reduce the quality of the captured images. In this work, we will focus on RF-based sensing as a complementary modality to vision and demonstrate the benefits of integrating it into the GS pipeline for efficient 3D rendering. Specifically, we aim to: (i) leverage a single radar transmission, providing only sparse RF-based depth measurements $\{y_t\}_{t=1}^T$ corresponding to locations $\{\mathbf{x}_t\}_{t=1}^T$, to efficiently predict depth $y_{\text{unobserved}}$ at any unobserved location $\mathbf{x}_{\text{unobserved}}$, addressing the RF-based depth-map reconstruction problem\footnote{The radar depth-map reconstruction problem considered here is analogous to the well-studied radio map reconstruction problem (e.g., \cite{polyzos2024bayesian}), which relies on received signal strength observations, but in our setting uses depth measurements obtained from radar transmissions.}; and (ii) use the resulting RF-driven depth map to construct a 3D PC as an alternative to the vision-based SfM or pre-trained models, and evaluate how this RF-informed initialization translates to GS-based rendering performance. 

The next section introduces the proposed RF-based depth-map reconstruction approach, which adapts conventional GPs through a localization scheme. In this framework, different spatial regions are modeled by distinct local GPs to improve computational efficiency and ensure that only the most relevant measurements influence each region, thereby yielding high-quality depth predictions with well-controlled uncertainty, as will be delineated next.

\section{RF-Driven Depth-Map Reconstruction for GS}

Given a single radar transmission that provides sparse RF-based depth measurements $\mathbf{y}_T = [y_1,\ldots,y_T]^\top$ at corresponding spatial locations $\mathbf{X}_T = [\mathbf{x}_1,\ldots,\mathbf{x}_T]^\top$, the objective is to learn a function $f(\cdot): \mathbf{x}_t \rightarrow y_t, \forall t$ such that the depth value $y_{\text{unobserved}}$ can be accurately predicted at any unobserved location $\mathbf{x}_\text{unobserved}$. In this work, we will capitalize on GPs as an efficient Bayesian modeling framework capable of learning an unknown function while simultaneously providing principled uncertainty estimates \cite{Rasmussen2006gaussian, lu2023surrogate}.

\subsection{Conventional GP-based learning}

GP-based learning begins by assuming that the unknown function $f$ is a random function endowed with a Gaussian prior over its evaluations; specifically, $\mathbf{f}_T := [f(\mathbf{x}_1),\ldots, f(\mathbf{x}_T)]^\top \sim \mathcal{N} ({\bf 0}_T, {\bf K}_T)$ where ${\bf K}_t$ is the covariance matrix whose $(m,m')$ entry is $[{\bf K}_t]_{m,m'} = {\rm cov} (f(\mathbf{x}_m), f(\mathbf{x}_{m'})):=\kappa(\mathbf{x}_m, \mathbf{x}_{m'})$, and $\kappa$ denotes the kernel function that assesses the pairwise similarity between distinct inputs $\mathbf{x}_m$ and $\mathbf{x}_{m'}$ \cite{Rasmussen2006gaussian}. The next assumption links the observed measurements $\mathbf{y}_T$ to the latent function values $\mathbf{f}_T$ through a factored batch conditional likelihood $p (\mathbf{y}_T| \mathbf{f}_T; \mathbf{X}_T)  \!=\! \prod_{t = 1}^{T} p(y_t| f(\mathbf{x}_{t}))$ where $ p(y_t| f(\mathbf{x}_{t})) = \mathcal{N}(f(\mathbf{x}_{t}), \sigma_{n}^2)$ since $y_t$ can be expressed as $y_t = f({\bf x}_t) + n_t$ with $ n_t \sim \mathcal{N}(0,\sigma_n^2)$ being Gaussian noise uncorrelated across $t$. 

With the GP prior and batch conditional likelihood at hand, the function posterior pdf of $f(\mathbf{x})$ at any unobserved location $\mathbf{x}$ can be computed through Bayes' rule as \cite{Rasmussen2006gaussian} \begin{align}
	p(f(\mathbf{x})|\mathbf{X}_T, \mathbf{y}_T) = \mathcal{N}(\mu_T (\mathbf{x}), \sigma_{T}^2(\mathbf{x})) 
	\label{eq:posteriorgaussian}
\end{align}
with mean and variance given in closed form as 
\begin{subequations}
\begin{align}	
\mu_T ({\bf x}) & = \mathbf{k}_T^{\top} ({\bf x}) (\mathbf{K}_T + \sigma_{n}^2
 \mathbf{I}_T)^{-1} \mathbf{y}_t \label{eq:mean}\\
\sigma_{T}^2 ({\bf x})& = \!\kappa(\mathbf{x},\mathbf{x})\! -\! \mathbf{k}_T^{\top} ({\bf x}) (\mathbf{K}_T\! +\! \sigma_{n}^2 \mathbf{I}_T)^{-1} \mathbf{k}_T ({\bf x}) \label{eq:variance}
\end{align}\label{eq:plain_gpp}
\end{subequations}
where $\mathbf{k}_T ({\bf x}) := [\kappa(\mathbf{x}_1, \mathbf{x}), \ldots, \kappa(\mathbf{x}_T,  \mathbf{x})]^\top$. Note that the posterior mean in \eqref{eq:mean} provides a point prediction of the depth value corresponding to the unobserved location $\mathbf{x}$, and the posterior variance in \eqref{eq:variance} quantifies the associated uncertainty. 

Although effective in diverse practical settings, conventional GP-based learning incurs $\mathcal{O}(T^3)$ complexity (c.f. \eqref{eq:plain_gpp}), making its adoption impractical as $T$ becomes large. Moreover, in the RF-based depth prediction setting considered here, depth measurements obtained from faraway locations have negligible influence on the depth at a specific point, as only spatially proximate observations are informative. Motivated by these insights, we next introduce a more computationally efficient, localization-based GP approach that yields well-calibrated uncertainty and more accurate depth predictions.

\subsection{Localized GPs for efficient depth prediction}

Rather than relying on a global GP model defined over the entire depth domain $\mathcal{R}$, we propose a principled localization strategy in which the space is partitioned into non-overlapping regions, $\mathcal{R} = \{r_1,\ldots,r_R\}$. For each region $r \in \{r_1,\ldots,r_R\}$ we instantiate a separate GP that conditions only on the observations $\{y_i^{(r)}\}_i^{T^{(r)}}$ associated with that region. Specifically, for any query location $\mathbf{x}$ within region $r$, the GP-posterior pdf $p(f(\mathbf{x})|\mathbf{X}_T^{(r)}, \mathbf{y}_T^{(r)}) = \mathcal{N}(\mu_T^{(r)} (\mathbf{x}), \sigma_{T}^{(r)2}(\mathbf{x}))$ is computed using \eqref{eq:plain_gpp} considering only the $T^{(r)}$ relevant region-specific data $\{\mathbf{X}_{T^{(r)}}, \mathbf{y}_{T^{(r)}}\}$.

The intuition behind this approach is threefold:
(i) depth measurements originating from distant regions are largely irrelevant and have negligible influence on predictions within the local region; (ii) each region processes only $T^{(r)}\ll T$ observations, reducing the GP computational complexity to $\mathcal{O}(T^{(r)3})$; and (iii) by restricting GP-modeling and inference to the most pertinent measurements, the localization strategy yields better-controlled posterior variance and more accurate depth predictions within each region. It is worth noting that, although the localization strategy introduces $R$ separate GP models, each model can be evaluated independently, enabling full parallelization and thereby preserving computational efficiency.

With the PC obtained from the proposed localized GP-based depth-map reconstruction approach, the GS Gaussian functions are initialized accordingly, and their parameters are subsequently optimized using the available training images.

\vspace{0.1cm}

\noindent \textbf{Remark.} Although the multimodal framework in this work focuses primarily on vision and RF-based sensing, it can naturally incorporate LiDAR measurements for PC prediction whenever available. Exploring alternative strategies for jointly leveraging all three sensing modalities, beyond PC construction, belongs to our future research agenda.

\begin{figure}[htbp]
    \centering
    \includegraphics[width=1.0\columnwidth]{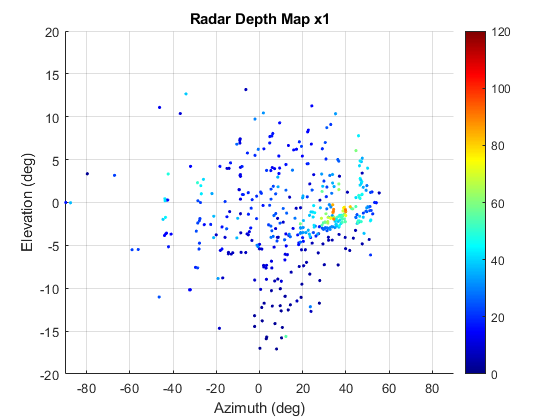}
    \caption{Initial depth map from a single radar transmission consisting of sparse depth measurements/observations.}
    \label{fig:Radar1measurement}
\end{figure}

\begin{figure*}[ht!]
\centering

\begin{subfigure}{0.32\textwidth}
  \centering
  \includegraphics[width=1.0\columnwidth]{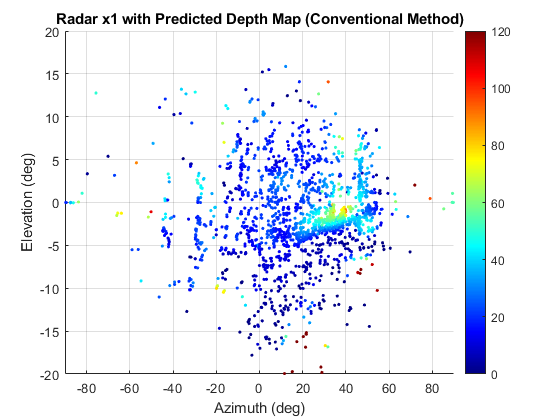} 
  \caption{Conventional GP prediction}
  \label{fig:conv}
\end{subfigure}
\hfill
\begin{subfigure}{0.32\textwidth}
  \centering
  \includegraphics[width=1.0\columnwidth]{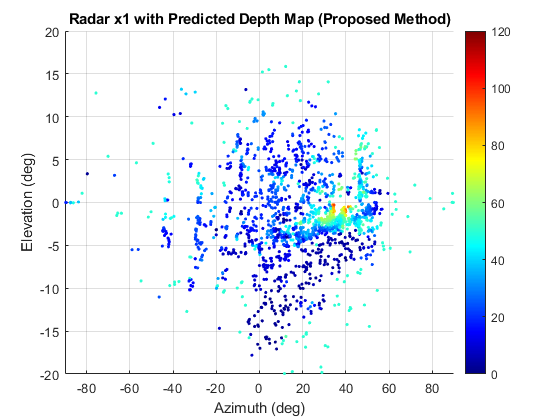}
  \caption{Proposed localized GP prediction}
  \label{fig:prop}
\end{subfigure}
\hfill
\begin{subfigure}{0.32\textwidth}
  \centering
  \includegraphics[width=1.0\columnwidth]{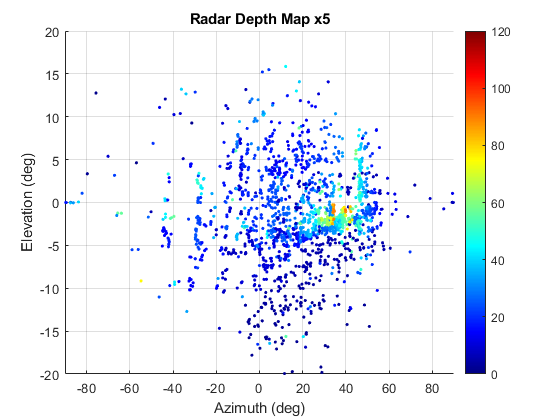}
  \caption{Ground truth}
  \label{fig:truth}
\end{subfigure}

\caption{Comparison of (a) the conventional `global' GP depth predictor and (b) the proposed localized GP predictor, shown alongside (c) the ground-truth depth map obtained from five radar transmissions.}
\label{fig:depth prediction mean}
\end{figure*}

\begin{figure}[htbp]
    \centering
    \includegraphics[width=1.0\columnwidth]{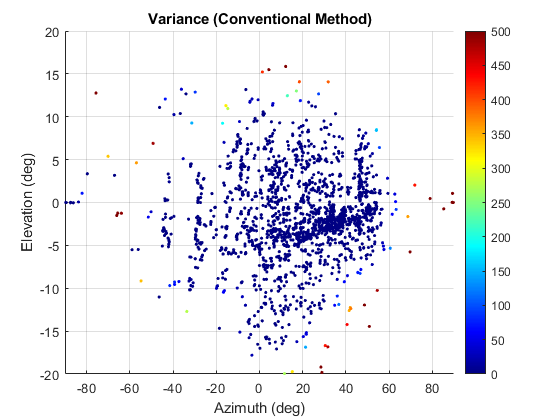}
    \caption{Depth variance at different angles using the conventional GP-based depth predictor.}
    \label{fig:conventional_gaussian_var}
\end{figure}

\begin{figure}[htbp]
    \centering
    \includegraphics[width=1.0\columnwidth]{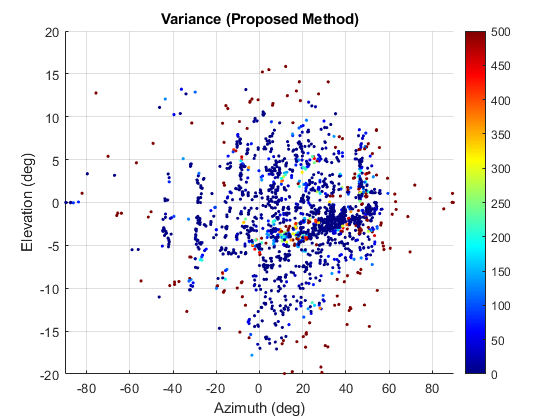}
    \caption{Depth variance at different angles using the proposed localized GP-based depth predictor.}
    \label{fig:proposed_gaussian_var}
\end{figure}

\begin{figure*}[!t]
    \centering
    \includegraphics[width=0.999\linewidth]{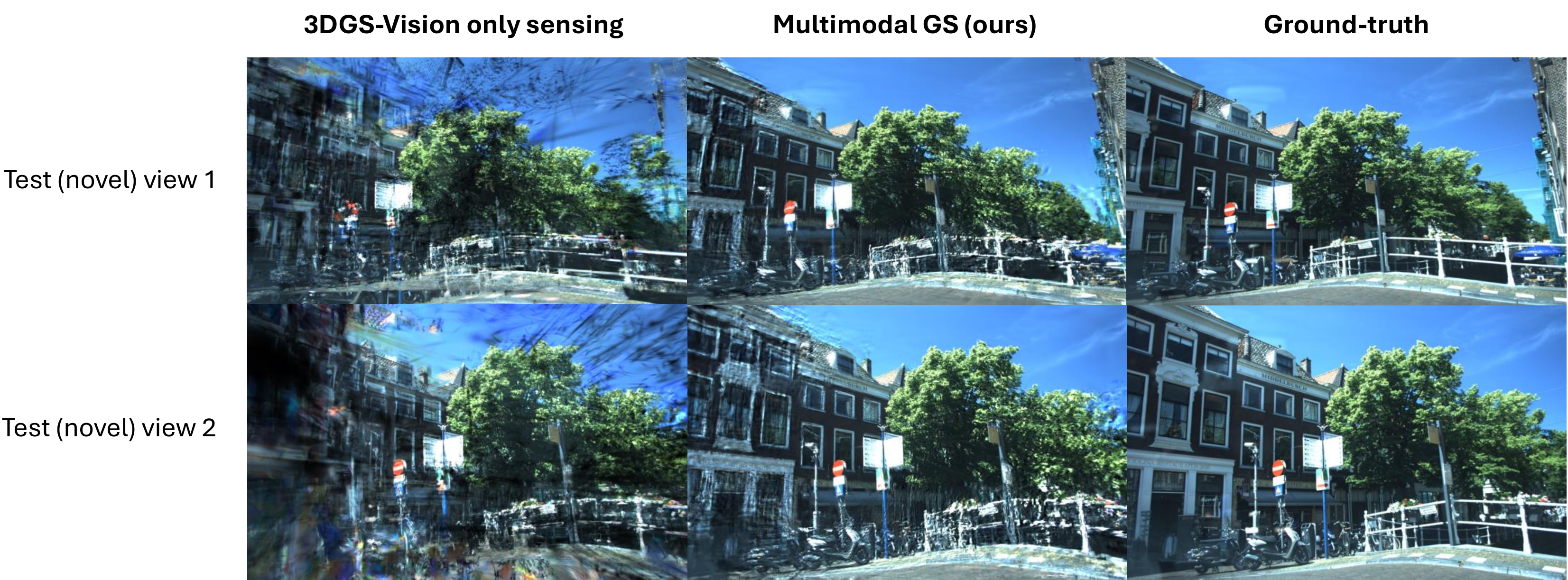}
    \caption{Visual comparison of the proposed multimodal GS with the conventional vision-only 3DGS on two indicative test (novel) viewpoints. It is evident that the rendered images produced by the proposed multimodal GS approach exhibit substantially improved quality compared to the conventional unimodal 3DGS baseline.}
    \label{fig:real_qual_plot}
\end{figure*}

\section{Numerical tests}

\subsection{Implementation details}

We evaluate the 3D rendering performance of the proposed approach on the View-of-Delft dataset \cite{View-of-Delft}, which contains urban driving scenes captured from a vehicle equipped with both camera and radar sensors. The part of the urban scene used in this study is captured by $M=35$ total images with $N_{\text{train}}=12$ allocated for training and the remaining $N_{\text{train}}=23$ used for testing. In addition to RGB images captured from camera sensors, each radar transmission produces a sparse depth map computed from returned echoes at specific azimuth and elevation angles, as determined by the radar signal processing pipeline. The dataset provides depth measurements for sparse points spanning azimuth angles in \([-90^\circ,\, 90^\circ]\) and elevation angles in \([-20^\circ,\, 20^\circ]\). This setting is particularly challenging for conventional GS-based rendering due to the very limited number of available images for GS initialization and training, combined with the confined scene coverage they provide. 

In our experimental setup for the proposed multimodal framework, the observed RF-based measurements consist of observed depth values obtained from a single radar transmission captured in the same time slot as the first training image. For the proposed localized GP-based approach for efficient depth-map (and corresponding PC) reconstruction from RF measurements, each regional GP model employs an RBF kernel $\kappa^{(r)}$ whose lengthscale hyperparameter is optimized by maximizing the marginal log-likelihood using only the depth observations within that region. With the PC obtained either from our proposed localized GP-based method or from traditional vision-only baselines, used for GS initialization, we follow the standard GS pipeline of \cite{kerbl20233d} for rendering. The GS model is trained using  $N_{\text{train}}=12$ images, optimizing the combined L1 and D-SSIM loss function as in \cite{kerbl20233d}, over 30000 training iterations. 

The localized GP-based approach for depth prediction was executed on an Intel Core i7-5930K CPU, while the GS training process was conducted on a Tesla V100-SXM2-16GB GPU hosted on Amazon servers. For GS training and rendering, we use the publicly available implementation at \url{https://github.com/graphdeco-inria/gaussian-splatting}. For the vision-only GS baseline, the PC used for GS initialization was constructed using COLMAP.



\subsection{Numerical results}

To demonstrate the merits of the proposed multimodal GP framework, we adopt a twofold evaluation strategy:
(i) we first assess the effectiveness of the localized GP approach in predicting unobserved depth values from sparse radar-based depth measurements/observations; and
(ii) we evaluate how the RF-derived PC produced by the localized GP method enhances GS rendering performance, while also reducing the processing cost for GS initialization compared to its vision-only GS-based counterpart.  



\vspace{0.1cm}

\subsubsection{Efficient RF-based depth prediction}

Starting with a sparse set of depth values obtained from a single radar transmission, depicted in Fig.~\ref{fig:Radar1measurement}, we use the proposed localized GP-based framework to predict the depth values at unobserved angles to obtain a more informative PC to initialize the Gaussian functions of GS. 
Using as ground truth the depth values obtained from five radar transmissions--available in the View-of-Delft dataset but treated as unknown during prediction and used solely for evaluation--we compare in Fig.~\ref{fig:depth prediction mean} the performance of the proposed localized GP approach against the conventional `global' GP predictor \cite{Rasmussen2006gaussian}. The results demonstrate that the depth map produced by the localized GP approach is closer to the ground truth, highlighting the advantages of the proposed localization strategy in terms of prediction accuracy.
Quantitatively, the conventional GP method yields an overall mean absolute error of 13.07 m, whereas the proposed localized strategy substantially reduces this error to 10.57 m. 

In addition to the predicted mean depth values, we also illustrate the variance of the conventional GP predictor and the proposed localized GP predictor in Figs. \ref{fig:conventional_gaussian_var} and \ref{fig:proposed_gaussian_var}, respectively. The results indicate that our proposed predictor provides a more detailed and spatially coherent representation of uncertainty, as the predicted depth variance adapts to local measurement characteristics. Lastly, we compare the localized approach with the conventional GP counterpart in terms of running time. Table~\ref{tab:computation_time} reports the runtime of the competing alternatives, where the proposed approach exhibits a substantially lower computational cost. This improvement is expected, as our method processes only the observations within each region rather than operating on the full set simultaneously, thereby reducing the overall computational complexity.  




\begin{table}[t]
    \centering
    \caption{Running time comparison for depth map reconstruction}
    \scalebox{1.05}{
    \begin{tabular}{lccc}
        \toprule
        \textbf{Method} & \textbf{Running time $\downarrow$}  &   \\
        \midrule
        Conventional GP-based prediction     & 9.39 s \\
        Localized GP-based prediction (ours)  & \textbf{0.81 s} &  \\
        \bottomrule
    \end{tabular}}
    \label{tab:computation_time}
\end{table}


\vspace{0.1cm}

\subsubsection{Gaussian splatting rendering performance}

Next, we demonstrate how the proposed multimodal GS framework, leveraging the RF-assisted PC generated by our localized GP method from a single radar transmission, can assist GS performance relative to conventional vision-only GS approaches. For our multimodal GS approach, note that a single radar transmission provides depth measurements only at a limited set of detected angles; it offers no information about depth at undetected angles or which of those angles contain objects. Therefore, unlike the previous subsection—where detected angles from five radar transmissions were used to validate the effectiveness of the proposed localized GP depth predictor—in practical scenarios such information is not available a priori. To mimic such a realistic practical setting, we instead generate random points within the azimuth and elevation ranges, predict their depth using our localized GP approach, and subsequently use the resulting PC for GS initialization. It is worth noting that, when constructing the PC, we retain only the depth estimates with lower posterior variance (those corresponding to higher confidence) ensuring a more reliable and accurate PC.  

In Table~\ref{tab:GS_perf}, we compare the rendering performance—using the widely adopted LPIPS, SSIM, and PSNR metrics—of the proposed multimodal GS approach, which leverages the RF-assisted PC for GS initialization and uses the available training images for GS training, against the conventional unimodal GS baseline, which relies solely on the training images for both initialization and training. It can be clearly seen that the multimodal GS approach achieves a markedly improved rendering performance compared to the vision-only GS baseline. Qualitatively, Fig.~\ref{fig:real_qual_plot} illustrates renderings from both approaches at two representative unseen (novel) test viewpoints, showing that the multimodal method produces outputs that more closely match the corresponding ground-truth images. This underscores the advantages of properly integrating RF-based and vision-based sensing modalities, compared to conventional unimodal vision-only GS approaches.

\vspace{0.1cm}

\noindent \textbf{Runtime comparison}. In contrast to conventional GS relying solely on training images and COLMAP to generate a PC, requiring \textit{4.43 mins} in our setting, the proposed radar-based depth predictor produces a complete PC from sparse depth measurements in approximately \textit{1 sec}, demonstrating a substantial improvement in computational efficiency for GS initialization.


\begin{table}[t]
    \centering
    \caption{LPIPS, SSIM and PSNR values for all competing methods in a certain scene of the View-of-Delft datset.}
    \scalebox{1.05}{
    \begin{tabular}{lccc}
        \toprule
        \textbf{Method} & \textbf{LPIPS} $\downarrow$ & \textbf{SSIM} $\uparrow$ & \textbf{PSNR} $\uparrow$ \\
        \midrule
        3DGS-Vision only sensing     & 0.5114 & 0.4161 & 13.339 \\
        Multimodal GS (ours)  & \textbf{0.4727} & \textbf{0.4628} & \textbf{15.032} \\
        \bottomrule
    \end{tabular}}
    \label{tab:GS_perf}
\end{table}

\section{Conclusions}
In this paper, we introduced a multimodal 3D scene rendering framework that integrates RF-based and visual sensing modalities to address key limitations of unimodal, vision-only GS pipelines. Leveraging sparse radar depth measurements, we developed a localized GP framework for efficient depth-map reconstruction that produces highly informative PCs with improved depth-prediction accuracy, better-calibrated uncertainty, and substantially reduced computational complexity compared to conventional `global' GP predictors. The resulting RF-driven PC is then used for GS initialization as an alternative to traditional vision-based pipelines. Numerical tests on a 3D scene from the View-of-Delft dataset demonstrated that (i) even a single radar transmission, when processed through the proposed localized GP approach, provides meaningful structural cues for 3D rendering; and (ii) RF-informed GS initialization achieves superior rendering fidelity compared to its vision-only counterpart, as evidenced by improvements in LPIPS, SSIM, and PSNR, while exhibiting reduced processing costs. These results highlight the strong potential of multimodal sensing, particularly the integration of RF and vision, for efficient, and high-quality 3D scene rendering.


\bibliographystyle{IEEEtranS}
\bibliography{my_bib}
\end{document}